\documentclass[conference]{IEEEtran}
\IEEEoverridecommandlockouts
\usepackage{cite}
\usepackage{amsmath,amssymb,amsfonts}
\usepackage{booktabs}
\usepackage{algorithmic}
\usepackage{graphicx}
\usepackage{textcomp}
\usepackage{xcolor}
\usepackage{hyperref}

\def\BibTeX{{\rm B\kern-.05em{\sc i\kern-.025em b}\kern-.08em
    T\kern-.1667em\lower.7ex\hbox{E}\kern-.125emX}}
\begin{document}

\title{Centrality of the Fingerprint Core Location
}

\author{\IEEEauthorblockN{1\textsuperscript{st} Laurenz Ruzicka}
\IEEEauthorblockA{\textit{DSS, SVS} \\
\textit{Austrian Institute of Technology}\\
Vienna, Austria \\
Laurenz.Ruzicka@ait.ac.at}
\and
\IEEEauthorblockN{2\textsuperscript{nd} Bernhard Strobl}
\IEEEauthorblockA{\textit{DSS, SVS} \\
\textit{Austrian Institute of Technology}\\
Vienna, Austria \\
Bernhard.Strobl@ait.ac.at}
\and
\IEEEauthorblockN{3\textsuperscript{rd} Bernhard Kohn}
\IEEEauthorblockA{\textit{DSS, SVS} \\
\textit{Austrian Institute of Technology}\\
Vienna, Austria \\
Bernhard.Kohn@ait.ac.at}
\and
\IEEEauthorblockN{4\textsuperscript{th} Clemens Heitzinger}
\IEEEauthorblockA{\textit{Institute of Analysis and Scientific Computing} \\
\textit{TU Wien}\\
Vienna, Austria \\
clemens.heitzinger@tuwien.ac.at}
}

\maketitle

\begin{abstract}
Fingerprints have long been recognized as a unique and reliable means of personal identification. Central to the analysis and enhancement of fingerprints is the concept of the fingerprint core. Although the location of the core is used in many applications, to the best of our knowledge, this study is the first to investigate the empirical distribution of the core over a large, combined dataset of rolled, as well as plain fingerprint recordings. We identify and investigate the extent of incomplete rolling during the rolled fingerprint acquisition and investigate the centrality of the core. After correcting for the incomplete rolling, we find that the core deviates from the fingerprint center by 5.7\% $\pm$ 5.2\% to 7.6\% $\pm$ 6.9\%, depending on the finger. Additionally, we find that the assumption of normal distribution of the core position of plain fingerprint recordings cannot be rejected, but for rolled ones it can. Therefore, we use a multi-step process to find the distribution of the rolled fingerprint recordings. The process consists of an Anderson-Darling normality test, the Bayesian Information Criterion to reduce the number of possible candidate distributions and finally a Generalized Monte Carlo goodness-of-fit procedure to find the best fitting distribution. We find the non-central Fischer distribution best describes the cores' horizontal positions. Finally, we investigate the correlation between mean core position offset and the NFIQ 2 score and find that the NFIQ 2 prefers rolled fingerprint recordings where the core sits slightly below the fingerprint center.
\end{abstract}


\section{Introduction}

Fingerprints have long been recognized as a unique and reliable means of personal identification, playing a pivotal role in forensic investigations, biometric authentication systems, and other applications requiring individual recognition. The distinct patterns formed by ridges and valleys on the fingertips are highly characteristic, making fingerprints a valuable tool in the field of forensic science.

Central to the analysis and enhancement of fingerprints is the concept of the fingerprint core. The core represents a key feature within the ridge structure, and is often assumed to be located near the center of the fingerprint \cite{manhua_fingerprint_2005}. Understanding the location of the core and its relationship to the finger position is of great importance for various purposes, such as template alignment \cite{manhua_fingerprint_2005}, finger type classification \cite{karu_fingerprint_1996}, and pose correction, where pose correction in the context of contact less fingerprint recordings describes the process of correcting the recording for different viewing angles of the finger \cite{tan_towards_2020, ruzicka_improving_2023, qijun_zhao_3d_2011}. 

Another application that requires empirical knowledge of the core distribution is the process of generating synthetic fingerprints \cite{noauthor_synthetic_2009}. The goal of generating a synthetic fingerprint is to mimic the natural properties of a real fingerprint to an extend that it is no longer possible to differentiate the two. Here, replicating the natural distribution of the core position could improve the generalization ability of neural networks trained on the synthetic data towards real data.

% The core of a fingerprint serves as a reference point for aligning and comparing different fingerprint images. By identifying the core's precise position, it becomes possible to align the ridges and valleys of multiple fingerprints, facilitating the accurate comparison and matching of fingerprints for identification purposes . This alignment process plays a critical role in forensic investigations, enabling the efficient search and retrieval of fingerprints from vast databases.

% In addition to alignment, the core is also utilized for classification purposes. Fingerprint classification systems often rely on the core's location as one of the key parameters for categorizing fingerprints into specific patterns, such as loops, whorls, and arches. By analyzing the core's position and its relationship to other ridge characteristics, classifiers can assign fingerprints to distinct classes, aiding in the organization and retrieval of fingerprint records.

% Moreover, the core's position within the fingerprint structure is relevant for pose correction. When contactless fingerprints are captured, the different rotations a finger can be presented to the sensor lead to potential distortions in the representation of the fingerprint structure. Understanding the deviation of the core from the center can provide insights into the extent of pose correction required, which can be crucial for accurately representing and reconstructing the original fingerprint structure.

Therefore, the research question addressed in this paper is: What is the distribution of the fingerprint core? To the best of our knowledge, this study is the first to analyse the empirical distribution of the fingerprint core on a large dataset of both rolled and plain fingerprint recordings. By investigating the distribution and therefore also the deviation of the core from the image center, we aim to gain a deeper understanding of the inherent structure of fingerprints. Additionally, we examine the linkage between incomplete rolling and deviations of the core from the segmented fingerprint image center to gain insights into the reliability of rolled fingerprint representations. Furthermore, we investigate the distribution of the cores' horizontal and vertical positions. Finally, we analyze the correlation between the NFIQ 2 score and the mean core offset.

In the following sections, we will present related work, our methodology, datasets used for analysis, core detection technique, and the results obtained. Also, we will discuss the implications of our findings and their significance for fingerprint analysis and identification. By shedding light on the deviation of the fingerprint core from the center, this study aims to contribute to the advancement of forensic science and biometric technologies.

\subsection{Related Work}

In the field of fingerprint analysis, there has been extensive research devoted to the definition and detection of the fingerprint core as well as to the classification of fingerprints. 

In the existing literature, two primary methods for defining the core point in fingerprints are prominent. The first method, proposed by Srinivasan and Murthy \cite{srinivasan_detection_1992} utilizes ridge orientation to determine the core point. According to their definition, the core point is identified as the location with the maximal ridge curvature. The second way of defining a core uses the notion of the innermost closed ridge loop and is described in \cite{bahgat_fast_2013} by Bahgat, Khalil, Abdel and Mashali as well as in ISO/IEC 19794-1:2011 3.33. They define the core as being the topmost point of the innermost ridge line. Note that there can also be two cores inside one fingerprint in the case of a fingerprint of the double loop type, which has two innermost closed ridge loops. Also, there can be none for the case of a fingerprint of the arch type \cite{ametefe_fingerprint_2022}.

Traditional methods for detecting the core point in fingerprints have relied on manual examination and visual analysis by forensic experts \cite{yager_fingerprint_2004}. These classical techniques involve careful observation of ridge patterns, curvature, and the arrangement of loops and deltas within the fingerprint. More modern approaches use the ridge orientation \cite{bahgat_fast_2013}, Poincare index \cite{anwar_modified_2008}, curvature analysis \cite{mun_koo_curvature-based_2001} and various other methods \cite{yager_fingerprint_2004}.

In addition to the detection of the core position within a fingerprint, it is crucial to consider the context of the fingerprint within the image. Fingerprint images may not always be centered or aligned uniformly, which necessitates image segmentation techniques to extract the relevant fingerprint region and make the core position relative to the fingerprint itself, rather than the image as a whole.

A variety of image segmentation approaches have been employed in fingerprint analysis to address this challenge. Tomaz, Candeias and Shahbazkia analyse the pixel color information in the ST space \cite{tomaz_fast_2004}, while Bazen and Gerez use custom pixel features to segment the fingerprint image \cite{bazen_segmentation_2001}. More recent work has used a deep learning approach to fingerprint segmentation. For example, Murshed, Kline, Bahmani, Hussain and Schuckers developed a Convolutional based neural network that is combined with a Region Proposal Network \cite{murshed_deep_2022}, while Grosz, Engelsma, Liu and Jain develop an U-Net based model \cite{grosz_c2cl_2021}. 

An established open-source tool for fingerprint segmentation is the \textit{nfseg} tool from the NBIS toolset \cite{noauthor_nist_2010}. This tool was developed by the National Institute of Standards and Technology (NIST) and provides a reliable, fast and accurate way of segmenting fingerprint images. 

\section{Methods}

\subsection{Core Detection}

 In this work, we follow the ISO core definition to locate the core positions and use the state-of-the-art, commercially available, automated fingerprint identification system IDKit from Innovatrix \cite{noauthor_idkit_nodate}. For detecting the core in this study, we require a minimal image region of 90x90 pixels.

%  \cite{basak_detection_2012, iwasokun_fingerprint_2014} to automatically detect the core point in fingerprints. 

\subsection{Fingerprint Segmentation \& Relative Core Offset}

To evaluate the stability of the core's centrality in a fingerprint, a multi-step approach is employed. The first step involves segmenting and cutting out the fingerprint, which is accomplished using the nfseg tool. Additionally, nfseg rotates the fingerprint image, such that the major symmetry axis is aligned with the y-axis. It requires images of at least 200x200 pixels.

Once the fingerprint image is appropriately segmented and the image is cropped to the fingerprint region, the core points are detected within the image. If multiple cores are detected, we proceed in two ways: \textit{a)} we calculate the mean position of the cores and use this mean for further calculations and \textit{b)}, every core is treated individually. To assess the position of the core within the fingerprint, its offset from the central point of the cropped image is measured. We call this distance core offset $o_{core}^x$ and $o_{core}^y$, where $x$ is the index for the horizontal component and $y$ the index for the vertical component.

To facilitate the evaluation of core positions in a standardized manner, a normalization process is employed. In order to increase readability, we depict analogous expressions for the y-axis in this section in parenthesis. The measured core offsets $o_{core}^x$ ($o_{core}^y$) are divided by half the width (height) of the cropped fingerprint $w_{cropped}$, resulting in a score that represents the core's centrality. This score ranges from -1 to 1, indicating the relative position of the core within the fingerprint. A score of -1 suggests that the core is located at the left (upper) edge of the fingerprint, while a score of 1 indicates its proximity to the right (lower) edge. We call this score the relative core offset $ro_{core}^x$ ($ro_{core}^y$), as shown in Equation \ref{equ:ro}. 

\begin{equation} \label{equ:ro}
ro_{core}^x = \frac{2 o_{core}^x}{w_{cropped}}
\end{equation}
and analogous for the y-coordinate.

In order to investigate the variability of the centrality of the core, we calculate the mean of the absolute values of the relative core offsets $aro_{core}^x$ ($aro_{core}^y$) for each finger position $f$, as shown in Equation \ref{equ:aro}.

\begin{equation} \label{equ:aro}
aro_{core}^x(f) = \frac{1}{N} \sum_{i}^{N} |ro_{core;i}^x|,
\end{equation}

where $N$ is the number of cores for a given finger position in the database and $i$ is the index of the core. The calculation for the y-coordinate is analogous.

\subsection{Incomplete Rolling}

While recording rolled fingerprints, we observed that the user often does not roll the finger completely from nail-to-nail, but misses the last part of the finger. This implies that the distance along the x-axis from the core in the collected fingerprint to the fingerprint boarders is biased towards the starting side of the rolling process. Especially for the case of operator assisted rolling, where an experienced operator controls the acquisition process and routinely rolls from a preferred side for a given left or right hand, a bias $b_{core}$ for the left vs right side can be expected. If there is no preferred rolling direction and the complete nail-to-nail rolling process was not enforced, the bias vanishes but larger relative core offset values are expected. Note that the x-axis is defined for the rotated and cropped fingerprint image and therefore aligns with the rolling direction and not the sensor recording surface.
This bias $b_{core}$ can be measured by the mean relative core offset, as seen in equation \ref{equ:b}.

\begin{equation} \label{equ:b}
b_{core} = \frac{1}{N} \sum_{i}^{N} ro_{core;i}^x
\end{equation}

A non vanishing relative core offset implies one of two things: Either the incomplete rolling is the cause for the bias or there is an anatomic tendency for a fingerprint core for a given finger to sit further on one side than on the other.

In this work, we assume that the core for a given participant is sampled from a distribution with zero mean, i.e. that the core is positioned evenly around the fingerprint center, and that therefore a non-vanishing bias is an indication for a statistical error induced by the incomplete rolling. We analyse this claim by comparing biases for plain and rolled fingerprints.

In order to correct for a non-vanishing bias, we calculate the corrected relative core offset $caro_{core}$ for each finger position $f$. For this calculation, we subtract the found bias, i.e. the mean relative core offset, from the calculated relative core offset, resulting in the corrected mean of absolve values of the relative core offsets $caro_{core}$, as shown in Equation \ref{equ:caro}.  

\begin{equation} \label{equ:caro}
caro_{core}(f) = \frac{1}{N} \sum_{i}^{N} |ro_{core;i}^x - b_{core}(f)| 
\end{equation}

\subsection{Distribution of Core Positions}

In order to determine the underlying distribution of the core positions for each finger, the initial step was assessing the normality of the data. The Anderson-Darling test \cite{anderson_asymptotic_1952} was specifically chosen as the preferred method due to its high power and sensitivity in detecting departures from normality, taking into account the entire distribution, including the tails, and assigning greater weights to extreme deviations. Should the assumption of normality not be met, an alternative approach was employed to identify a suitable set of distributions that best describe the data. This was achieved by employing the Bayesian Information Criterion (BIC) \cite{schwarz_estimating_1978}, which combines the complexity of a model with its performance into one score. It chose from a set of over 110 distributions and selected the overall best matches. In the next step, a Generalized Monte Carlo goodness-of-fit procedure \cite{noauthor_scipy_nodate} was used to find the best distribution of the previously selected set of distributions for each finger position.  
This procedure is often described as the parametric bootstrap test in literature \cite{stute_bootstrap_1993, genest_validity_2008, kojadinovic_goodness--fit_2012}. %Traditionally, the goodness-of-fit tests relied on pre-calculated critical values based on fixed significance levels. The Generalized Monte Carlo goodness-of-fit procedure however allows for performing the Monte Carlo trials adapted to their particular data.

%and it works as follows. 
%
% The estimation of unknown parameters in the specified distribution family is conducted using maximum likelihood estimation. The estimated parameters define the null-hypothesized distribution, representing the distribution from which the data were assumed to be sampled under the null hypothesis. We calculate the Anderson-Darling statistic for the given finger-wise $ro_ {core}$. Subsequently, numerous new samples, each comprising of the same number of observations as the original data, were drawn from the null-hypothesized distribution. For each drawn sample, unknown parameters were re-estimated, and the corresponding Anderson-Darling statistic was calculated between the sample and its fitted distribution. These computed statistic values collectively formed the Monte Carlo null distribution, distinct from the aforementioned null-hypothesized distribution. The p-value of the goodness-of-fit test is determined as the proportion of statistic values in the Monte Carlo null distribution that are as extreme as or more extreme than the statistic value computed for the provided data. Specifically, the p-value is calculated as the ratio of the number of statistic values in the Monte Carlo null distribution greater than or equal to the statistic value of the data plus one to the total number of elements in the Monte Carlo null distribution also plus one. \cite{lilliefors_kolmogorov-smirnov_1967}

\subsection{NFIQ 2 Evaluation}

The NIST Fingerprint Image Quality (NFIQ) Version 2 is a software tool designed for assessing the quality of fingerprint images in the context of biometric identification and verification systems \cite{tabassi_nist_2021}. It links image quality of optical and ink 500 DPI fingerprints to operational recognition performance. It works by quantifying 14 image quality attributes, including ridge valley uniformity, ridge flow continuity, and Minutiae quality. A higher NFIQ 2 score denotes superior image quality, whereas lower scores indicate lower-quality fingerprint images.

To assess the relationship between the NFIQ 2 score and the core offset in both the X and Y directions, we employed Spearman's correlation coefficient \cite{spearman_proof_1904}. Spearman's correlation coefficient is a statistic that quantifies the strength and direction of a monotonic relationship between two variables. It is particularly useful when the relationship between variables may not be linear and requires a measure of ordinal correlation.

% Spearman's correlation coefficient values tend to be close to 1 for samples with a strongly positive ordinal correlation, close to -1 for samples with a strongly negative ordinal correlation, and close to zero for samples with weak ordinal correlation. In our study, this statistic serves as a valuable tool to assess the association between the NFIQ 2 scores and the core offset.

In our analysis, the null hypothesis (H0) posits that there is no significant correlation between the NFIQ 2 scores and the core offset. The alternative hypothesis (Ha) suggests that there exists a significant correlation between the NFIQ 2 scores and the core offset. We investigate the hypothesis separately for horizontal and vertical core offset. Additionally, we also investigate the correlation between the $aro_{core}^x$, $aro_{core}^y$ and the NFIQ 2.

For the hypothesis test, the H0 can be transformed to a Student's t distribution with the number of samples minus two as the degrees of freedom \cite[p. 280]{howell_statistical_2013}. Note that this is only accurate for approximately 500 or more observations \cite{community_scipystatsspearmanr_nodate}. We quantified the comparison by computing the p-value, which represents the proportion of values in the null distribution that are as extreme as or more extreme than the observed value of Spearman's correlation coefficient. In a two-sided test, both elements of the null distribution greater than the transformed statistic and elements less than the negative of the observed statistic are considered "more extreme".

A small p-value indicates a low probability of obtaining such an extreme statistic value under the assumption that the NFIQ 2 scores and core offset are independent. In this context, a small p-value would indicate a significant correlation between the NFIQ 2 scores and core offset.

To determine the statistical significance of our findings, we set the significance level ($\alpha$) for the hypothesis test to $10^{-3}$. This significance level represents the threshold beyond which we consider the evidence against the null hypothesis to be compelling.

\subsection{Datasets}

This study uses 6 different datasets to investigate the centrality of the core. Some of the datasets provide a finger position description for each of the recorded fingerprints, which we denote in the FGP scheme introduced by NIST \cite[p. 18]{national_institute_of_standards_and_technology_american_2000}:
\begin{table}[tbhp]
\begin{center}
\caption{FGP values mapped to finger names.}
    \begin{tabular}{l|cc}
        Finger & Right & Left \\
        \toprule
        Thumb  &  1 &  6 \\
        Index  & 2 & 7 \\
        Middle & 3 & 8 \\
        Ring   & 4 & 9  \\
        Little & 5 & 10 \\
        \midrule
        Plain Thumb & 11 & 12 \\
        Unknown & 0 & 0 \\
    \end{tabular}
    
    \label{tab:fgp_values}
    \end{center}
\end{table}

\begin{table}[htb]
    \begin{center}
    \caption{Number of rolled fingerprints. }
    \begin{tabular}{c|cccc|c}
    \toprule
        FGP & AIT  & NIST 300a & NIST 302a & NIST 302b & + \\
        \midrule
        1   & 1083 & 815   & 935 & 521 & 3354 \\
        2   & 875 & 801   & 981 & 542 & 3199 \\
        3   & 528 & 819   & 981 & 605 & 2933 \\
        4   & 523 & 832   & 984 & 632 & 2971 \\
        5   & 941 & 803   & 952 & 616 & 3312 \\
        \hline
        6  & 1077  & 797   & 889 & 505 & 3268 \\
        7  & 512   & 806   & 963 & 575 & 2856 \\
        8  & 522  & 813   & 942 & 609 & 2886 \\
        9  & 525  & 816   & 993 & 630 & 2964 \\
        10 & 525  & 791   & 940 & 603 & 2859 \\
        \hline
        +  & 7111 & 8193 & 9560 & 5829 & 30602 \\
        \bottomrule
    \end{tabular}
    \label{tab:number_rolled_fingerprints}
    \end{center}
\end{table}

\begin{table}[htb]
    \begin{center}
    \caption{Number of plain fingerprints.}
    \begin{tabular}{c|cccc|c}
    \toprule
        FGP & NIST 300a & NIST 302b & Neurotechnology & PolyU & + \\
        \midrule
        0  & 0   & 0  & 840 & 2776 & 3616\\
        \hline
        1  & 0   & 226 & 0 & 0 & 226 \\
        2  & 704 & 228 & 0 & 0 & 932 \\
        3  & 662 & 255 & 0 & 0 & 917 \\
        4  & 768 & 270 & 0 & 0 & 1038 \\
        5  & 653 & 244 & 0 & 0 & 897 \\
        11 & 779 & 11  & 0 & 0 & 790 \\
        \hline
        6  & 0   & 236 & 0 & 0 & 236 \\
        7  & 771 & 236 & 0 & 0 & 907\\
        8  & 650 & 234 & 0 & 0 & 884\\
        9  & 779 & 257 & 0 & 0 & 936\\
        10 & 675 & 231 & 0 & 0 & 906\\
        12 & 748 & 151 & 0 & 0 & 899\\
        \hline
        + & 7189 & 2579 & 840 & 2776 & 9568 \\
        \bottomrule
    \end{tabular}
    \label{tab:number_plain_fingerprints}
    \end{center}
\end{table}

\subsubsection{AIT Dataset}
Weissenfeld, Schmid, Kohn, Strobl and Domínguez \cite{weissenfeld_case_2022} utilized data that was acquired by police officers. The data acquisition took place at a national police refugee registration center, ensuring a real-world setting for the collection process. The dataset comprises data acquired using two different devices: a contact-based device and a contactless device and for this study, only the contact-based part is used. The law enforcement agency utilized the IDEMIA TP 5300 scanner with 1000 dots-per-inch (DPI) for the rolled recordings and with 500 DPI for the plain slap-segmented ones. A total of 567 individuals submitted contact-based recordings.

The contact-based fingerprints were obtained as rolled fingerprints, with 20 prints per person corresponding to two recordings for each finger, as well as plain fingerprints. The capturing process involved police officers guiding the participants fingers.

Table \ref{tab:number_rolled_fingerprints} shows the number of fingerprints with at least one core that fit the nfseg image size constraints and well as the IDKit image region constraints.

\subsubsection{NIST Special Dataset 300a}
In collaboration with the FBI, the NIST has undertaken the digitization of 888 inked fingerprint arrest cards. The digitization process involved the use of either an Epson Perfection 4990 or an Epson V700 scanner, with a scan quality of 500 DPI. Finger positions are available for each recording. \cite{fiumara_nist_2018}

In this dataset, we have rolled fingerprint recordings as well as slap-segmented plain ones. Table \ref{tab:number_rolled_fingerprints} shows the number of rolled fingerprints after cleaning and Table \ref{tab:number_plain_fingerprints} shows the number of plain fingerprints after cleaning.

\subsubsection{NIST Special Dataset 302a \& 302b}
In September 2017, the Intelligence Advanced Research Projects Activity conducted the Nail to Nail (N2N) Fingerprint Challenge. The purpose was to obtain images of the complete surface area of a fingerprint, equivalent to a rolled fingerprint, without the need for a trained operator and from an unacclimated user. Participating Challengers utilized specially designed devices to capture these images. Alongside this, traditional operator-assisted live-scan rolled fingerprints were also collected. The collection of images resulting from the N2N Fingerprint Challenge is known as the Special Database 302. \cite{fiumara_nist_2021}

The Special Database 302a contains the unassisted recordings from different challenger devices and the Special Dataset 302b contains the operator-assisted live-scan rolled fingerprints. Both expose the finger position for each of the recordings. For the 302a dataset, only rolled recordings exist and Table \ref{tab:number_rolled_fingerprints} provides a more detailed mapping of the number of existing fingerprint recordings per finger after cleaning.

For the 302b dataset, both rolled and plain fingerprint recordings exist. Tables \ref{tab:number_rolled_fingerprints} and \ref{tab:number_plain_fingerprints} show the number of available fingerprint recordings after cleaning.

\subsubsection{Neurotechnology}
Neurotechnology provides two sample fingerprint databases on their website \cite{noauthor_download_nodate}. They are created using a \textit{Cross Matcher Verifier 300} and a \textit{DigitalPersona U.are.U}, both recording in 500 DPI. 360 + 480 recordings with one or two cores were available. No finger position information was provided. 

\subsubsection{PolyU Contact-based}
For this work, we used the contact-based subset of the database published by Lin and Kumar \cite{lin_matching_2018}. Similar to the Neurotechnology datasets, no finger information is available. The dataset consists of 2976 contact-based fingerprints obtained from 300 unique individuals. For 200 images, we did not find a core. The fingerprints were obtained using the URU 4000 fingerprint reader, the resolution of the sensor are 500 DPI.

\subsubsection{Rolled Combination}
For the combined results of the rolled fingerprints, we used the datasets: AIT Dataset, NIST 300a (rolled images only), 302a and 302b (rolled images only). Since we use the normalized core offset for our calculations, we were able to combine datasets in various resolutions. Table \ref{tab:number_rolled_fingerprints} shows the number of fingerprint recordings for each finger position used. In total, we had 30602 rolled fingerprint recordings.

\subsubsection{Plain Combination}
For the combined results of the plain fingerprints we used the following datasets: NIST Special Dataset 300a (plain images only), NIST Special Dataset 302b (slap-segmented images only), Neurotechnology and PolyU Contact-based. A finger-wise overview of the number of recordings per finger can be seen in table \ref{tab:number_plain_fingerprints}. In total, we had 9568 plain fingerprint recordings.

\section{Results \& Discussion}
\subsection{Bias \& Incomplete Rolling}

For the rolled fingerprints, we observed the following bias values, described as percentages of the $ro_{core}^x$ score in Table \ref{tab:bias_rolled}. The score can go from -100\% (left edge) to 100\% (right edge).  We find that the right hand with fingers 1-5 shows a positive bias, i.e. a tendency of the core to sit closer to the right side. This holds for both core handling cases, where results for \textit{b)} are shown in parenthesis in Table \ref{tab:bias_rolled}. This implies that we measured incomplete rolling where the finger was rolled starting with the right finger-nail edge. This effect was especially strong for middle, ring and little finger, with the strongest effect on the ring finger. One explanation for this could be obstruction of the non-rolled fingers in the rolling process. Another explanation could be the inter-connectivity of the tendons passing the carpal tunnel as well as the connection of the flexor digitorum superficialis to all fingers except the thumb. This makes it challenging to lift only the ring finger and to a lesser extend also the middle finger for most people. Therefore, the effort of separating the ring or middle finger from the other fingers is increased, which could lead to a faster termination of the rolling process.

\begin{table}[htb]
    \begin{center}
    \caption{Rolled Fingerprint Bias for Average (Both) Core X-Positions}
    \begin{tabular}{cc|cc}
    \toprule
        FGP & Bias [\%]    & FGP & Bias [\%] \\
        \midrule
        1   & 2.90 (4.10) & 6  & 0.42 (-0.78) \\
        2   & 0.67 (1.14) & 7  & 0.33 (-0.17) \\
        3   & 3.27 (4.31) & 8  & -1.11 (-1.92) \\
        4   & 5.17 (5.33) & 9  & -3.62 (-3.77) \\
        5   & 4.89 (5.67) & 10 & -0.50 (-0.78) \\
        
    \end{tabular}
    \label{tab:bias_rolled}
    \end{center}
\end{table}

\begin{table}[htb]
    \begin{center}
    \caption{Mean Core X-Position Plain Recordings for Average (Both) Core Positions}
    \begin{tabular}{cc|cc}
    \toprule
        FGP & Offset [\%]    & FGP & Offset [\%] \\
        \midrule
        1   & -1.50 (-0.15) & 6  & 2.35  (0.36) \\
        2   & -6.43 (-6.43) & 7  & 11.69 (10.51) \\
        3   & -5.67 (-5.83) & 8  & 9.79  (9.16) \\
        4   & 10.01 (9.98)  & 9  & -1.64 (-2.10) \\
        5   & -1.53 (-0.76) & 10 & 15.49 (14.06) \\
        \midrule
        11  & 7.21 (5.72)   & 12 & -0.72 (-1.50) \\
        0   & 2.32 (1.69)   & - & - \\
       
    \end{tabular}
    \label{tab:bias_plain}
    \end{center}
\end{table}

On the other side, the left hand also shows the pattern that the ring finger and the middle finger create the largest bias. The same reasoning of the hand anatomy, leading to the observed bias, could apply here. Also note the sign flip, indicating that the preferred starting edge of the rolling process changed. In future work, the effect of inverted roll directions and left versus right handiness on the bias could be explored with a dedicated dataset.

Additionally, also the plain fingerprint recordings showed a non-vanishing mean of the core x-position, similar to the bias of the rolled fingerprints. The magnitude of this can be seen in Table \ref{tab:bias_plain}. The same parenthesis notation as above applies here.

One observation of the mean fingerprint core in the plain recording setting is the mirrored mean core x-position shift. The sign of the offset of the core is flipped when comparing the left to the right hand. This could be an indication that the hand is placed tilted on the sensor, i.e. rotated along the finger axis, where the tilt between left and right hand is inverted. 

Another key observation is the comparison of the bias of the rolled fingerprints and the mean core position for plain fingerprint recordings. The tendency of the core in rolled fingerprints of the right hand was to sit closer to the right edge, while for plain recordings, only the thumb and the ring finger cores had the tendency to sit closer to the right edge. Also for the left hand, only the thumb and index finger cores both in plain and rolled fingerprints tend towards the right side of the finger. This is an indication that the reason for the non-zero core position offset is not an effect of human finger anatomy, but rather of the recording process. 

Finally, also the y-position of the mean fingerprint core offset can be calculated. For plain/rolled fingerprint recordings, we find the following offsets from the image center in percentages of the total $ro_{core}^y$, -100\% depicts the upper finger edge and 100\% the lower finger edge. The same parenthesis notation for core handling as above is used: 3.37\%/15.94\% (1.65\%/15.59) for right thumbs, -8.39\%/-5.28\% (-9.10\%/-5.12\%) for right index fingers, -16.80\%/-6.93\%  (-16.84\%/-6.34\%) for right middle fingers, -14.70\%/-8.36\% (-14.27\%/-7.35\%) for right ring fingers, -13.10\%/-13.33\% (-12.77\%/-11.62\%) for right little fingers, 5.15\%/15.03\% (2.61\%/14.45\%) for left thumbs, -9.66\%/-6.83\% (-9.75\%/-6.63\%) for left index fingers, -19.75\%/-8.37\% (-19.95\%/-8.48\%) for left ring fingers and -12.60\%/-13.15\% (-12.24\%/-11.65) for left little fingers. For the unknown finger position we found an offset of 5.18\%/- (4.76\%/-).

The direction of the y-offset of the mean core position agrees for all 10 fingers between plain and rolled recordings. This could be due to the fact that plain recordings in the observed dataset are taken from the same participants as the rolled recordings, with the exception of the PolyU and Neurotechnology datasets, which had no finger information.

\subsection{CARO \& ARO}

The $caro_{core}$ from equation \ref{equ:caro} is an important measure for how strongly the core is scattered around the bias position. It is the central indication of how stable the centrality of the core position is. Its non-bias corrected counterpart, the $aro_{core}$ from equation \ref{equ:aro} is an important measurement for the expected variability in a real recording environment, where incomplete rolling and therefore a non-vanishing bias is to be expected.

We found the following $caro_{core}$ scores for plain/rolled recordings using the core handling method \textit{a)}, which are written as percentages of the $ro_{core}$ score. We use core handling method \textit{a)}, because we want a single scalar per image to quantize the centrality of the core position as a landmark. This score can vary from 0\% (no deviation, every core in center) to 100\% (maximal deviation, every core at the finger edge). Additionally, we report the $aro_{core}$ scores for comparison in parenthesis. This can be seen in Table \ref{tab:caro_aro}. 

\begin{table}[htb]
    \begin{center}
    \caption{CARO (ARO) Scores for Rolled Fingerprints}
    \begin{tabular}{cc|cc}
    \toprule
        FGP & CARO (ARO) [\%]    & FGP & CARO (ARO) [\%] \\
        \midrule
        1   & 6.7 $\pm$ 5.6 (7.4 $\pm$ 5.9) & 6  & 7.2 $\pm$ 6.2 (7.7 $\pm$ 6.4) \\
        2   & 7.6 $\pm$ 6.9 (8.5 $\pm$ 7.1) & 7  & 7.0 $\pm$ 6.0 (7.7 $\pm$ 6.3) \\
        3   & 6.6 $\pm$ 5.7(7.3 $\pm$ 6.0)  & 8  & 7.0 $\pm$ 5.7 (7.4 $\pm$ 6.0) \\
        4   & 5.7 $\pm$ 5.2 (6. 6$\pm$ 5.6) & 9  & 6.3 $\pm$ 5.5 (6.9 $\pm$ 5.8) \\
        5   & 6.4 $\pm$ 5.3 (7.2 $\pm$ 5.5) & 10 & 6.6 $\pm$ 5.4 (7.2 $\pm$ 5.7) \\
        \bottomrule
    \end{tabular}
    \label{tab:caro_aro}
    \end{center}
\end{table}

One key remark is that the scattering of the core around the fingerprint center is similar between all fingers and falls within a range of 5.7\% $\pm$ 5.2\% to 7.6\% $\pm$ 6.9 for the different fingers. This quantifies the centrality of the core as a landmark for various fingerprint enhancement techniques, such as pose correction for contactless fingerprint recordings, where the centrality of the core is the assumption for calculating the viewing angle.  

For the $aro_{core}^y$ for rolled fingerprints, we found scores from 9\% $\pm$ 6\% for the right index finger to 12\% $\pm$ 8\% for the right thumb. Both thumbs had the highest values and the index fingers the lowest. The other fingers were all in the range of around 10\%. This indicates that the core varies in height roughly equally for all fingers, with a slight increase for the thumbs and decrease for the index fingers.

\subsection{Distribution of the Core}

The Anderson-Darling test revealed that the assumption of normality of the $caro_{core}$ data could be rejected with a significance level of under $10^{-16}$ for all rolled fingerprints. For the plain recordings, normality of $caro_{core}^x$ could only be rejected for the unlabeled fingers with a FGP of 0 and the $caro_{core}^y$ for all fingers except FGP 12. Therefore, we continued the testing with the BIC for the rolled fingerprint recordings in both axis, as well as the vertical core position of the plain fingerprints. 

The overall best BIC scores could be achieved with the Logistic distribution. Other well performing distributions that were added to the set of candidate distributions are: Laplace \cite[p. 930]{abramowitz_handbook_1972}, Cauchy \cite[p. 930]{abramowitz_handbook_1972}, Dagum (also known as Mielke) \cite{dagum_study_1999}, non-central Fischer (NCF) \cite{cabana_f_2011, ramirez_noncentral_2004}, Burr \cite{burr_cumulative_1942}, Normal and the Lognormal distribution \cite{galton_xii_1997}.

Of the eight candidate distributions, Logistic,  Laplace and Cauchy distributions could be rejected with a significance level of 5\% for all fingers. Both Dagum and NCF distribution achieved high p-values of up to 0.94 for the NCF and 0.70 for Dagum.

\begin{table}[htb]
    \begin{center}
    \caption{Best fitting distributions - Dagum (Dag), Lognorm (LNo), Burr (Bur), Norm (Nor)}
    \begin{tabular}{c|cc|cc|cc|cc}
    \toprule
        -  & \multicolumn{4}{c|}{Plain} & \multicolumn{4}{c}{Rolled} \\
        \hline
         - & \multicolumn{2}{c|}{x} & \multicolumn{2}{c|}{y} & \multicolumn{2}{c|}{x} & \multicolumn{2}{c}{y}\\
        FGP & Dist. & p [\%] & Dist. & p [\%]& Dist. & p [\%]& Dist. & p [\%] \\
        \midrule
        0   & Dag & 70 & NCF & 78 & - & - & - & - \\
        1   & LNo & 67 & NCF & 20 & Bur & 11 & NCF & 95 \\
        2   & LNo & 56 & Bur & 60 & NCF & 32 & NCF & 71 \\
        3   & Bur & 47 & NCF & 80 & NCF & 37 & Bur & 37 \\
        4   & NCF & 94  & NCF & 55 & NCF & 85 & Bur & 29 \\
        5   & NCF & 50 & NCF & 73 & NCF & 84 & Bur & 42 \\
        6   & NCF & 84 & NCF & 44 & NCF & 65 & Bur & 36 \\
        7   & NCF & 56 & Bur & 55 & NCF & 73 & NCF & 74 \\
        8   & NCF & 28 & NCF & 54 & NCF & 33 & NCF & 30 \\
        9   & NCF & 67 & Bur & 31 & NCF & 60 & NCF & 78 \\
        10  & Nor & 75 & NCF & 67 & Bur & 41 & NCF & 61 \\
        11  & NCF & 33 & Bur & 43 & - & - & - & - \\
        12  & Nor & 72 & NCF & 23 & - & - & - & - \\
        \bottomrule
    \end{tabular}
    \label{tab:distributions_p_values}
    \end{center}
\end{table}

The results for all the fingers can be seen in Table \ref{tab:distributions_p_values}.In most cases, the NCF distribution performed best. For rolled fingerprints, all fingers could best be described with either the NCF or the Burr distribution. For both plain as well as rolled fingerprint recordings, all cases where the NCF distribution did not have the highest goodness-of-fit p-value, it was the runner up with the second highest p-value.

We therefore conclude that the NCF is best suited to describe the core position and suggest to implement those for the generation of synthetic fingerprint samples.

\begin{figure}[htbp]
    \centerline{\includegraphics[width=0.8\linewidth]{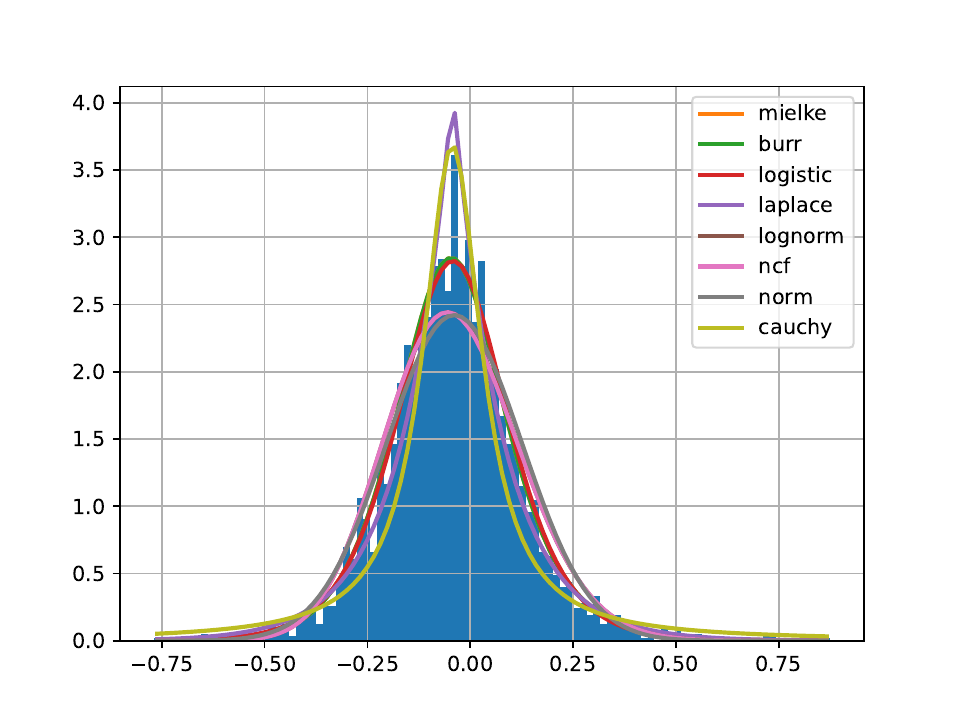}}
    \caption{Fitted distributions for the left ring finger (FGP 9).}
    \label{fig:dist}
\end{figure}

Exemplary, the chosen distributions fitted for the left ring finger can be seen in Figure \ref{fig:dist}. 

\subsection{NFIQ 2 Correlation }

\begin{figure}[htbp]
    \centerline{\includegraphics[width=\linewidth]{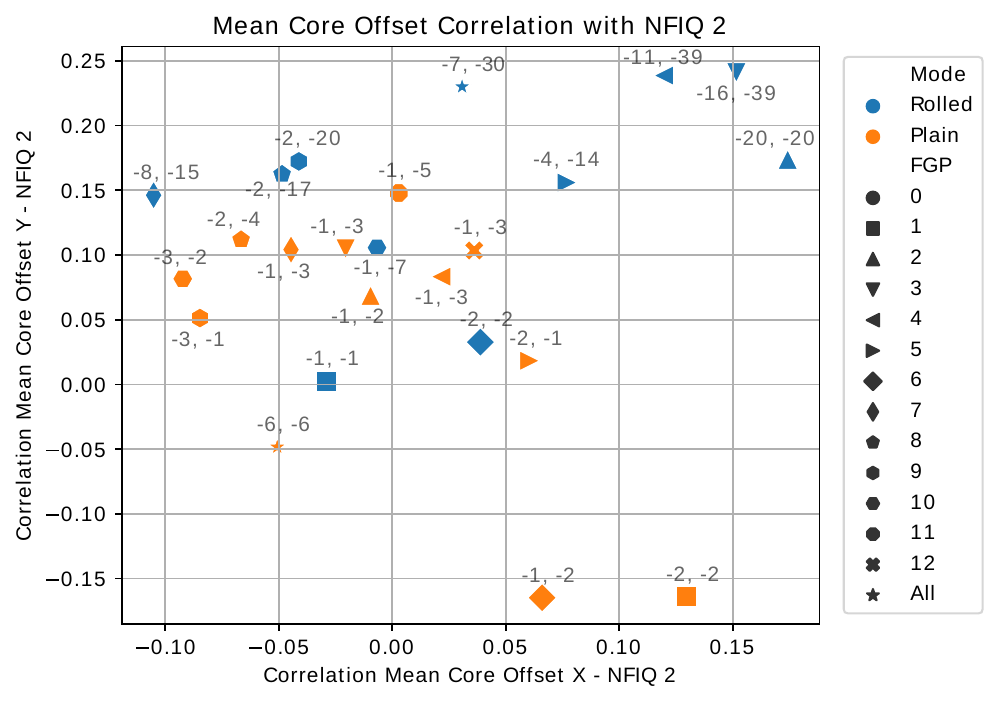}}
    \caption{Correlations between the Mean Core Offset in X and Y direction and the NFIQ 2 score for rolled (blue) and plain (orange) fingerprint recordings, for each finger position (see markers in legend).}
    \label{fig:correlations}
\end{figure}

In this section, we present the results of our analysis, which aimed to investigate the correlation between the mean $ro_{core}^x$, mean $ro_{core}^y$ and the NFIQ 2 scores, as well as the correlation between both $aro_{core}$ scores and the NFIQ 2 scores.

For rolled fingerprint recordings, we found a significant, positive correlation of 0.23 between the mean $ro_{core}^y$ and the NFIQ 2 scores. For the x component of the mean core offsets of the rolled fingerprint recordings, as well as both components of the mean core offsets of plain fingerprint recordings, no indicative correlation with a magnitude of over 0.06 could be found.

Therefore, we conclude that for rolled fingerprint recordings, a non-zero core offset in the vertical fingerprint direction where the core is shifted to the lower part of the image is favoured by the NFIQ 2 score. Furthermore, we observed a very small correlation of -0.09 between the $aro_{core}^y$ and the NFIQ 2 score. Therefore, there seems to be a preference of the NFIQ 2 score for core locations on the lower halve of the image, with a tendency to prefer cores closer to the center.

Interestingly, we did not observe a relevant correlation in x direction between the mean core offset and the NFIQ 2 score. Also the correlation of the $aro_{core}^x$ is not statistically significant. This implies that there is either no preferred side by the NFIQ 2 score or that the horizontal component of the core position does not strongly influence the NFIQ 2 score. 

A more detailed analysis on a finger-wise basis can be seen in Figure \ref{fig:correlations}. Here, the correlation results for plain and rolled fingerprint recordings for each finger are depicted. The labels are the exponents of the p-values of the Spearman's correlation test with 10 as their basis. The first number describes the p value of the correlation in x direction and the second in y direction. 

Notable, with the exception of both thumbs in plain recordings, the correlation between the mean $ro_{core}^y$ and the NFIQ 2 score is close to zero or positive. And for the thumbs in plain recordings, the results are not statistically significant with their p values of above $10^{-3}$. More data is required to come to a conclusion regarding the thumbs in plain fingerprint recordings.

\section{Conclusion}

We found a non-vanishing bias in our combined rolled dataset. This effect was especially strong for middle (right 3\%, left -1\%), ring (right 5\%, left -4\%) and little (right 5\%, left -1 \%) finger, with the strongest effect on the ring finger. 
Furthermore, we also observed a non-vanishing bias for the plain recording setting, but comparing rolled with plain recordings, we found that the preferred side is not linked to the finger position value. This indicates that the fingerprint core position does not have an anatomically preferred side but rather that the non-vanishing bias originates from incomplete rolling in the case of rolled fingerprint recordings and placing that hand tilted on the sensor for the plain fingerprint recordings. 

In addition to this finding, we measured the variability of the core position, given as $caro_{core}^x$ and $aro_{core}^y$. We found the $caro_{core}^x$ to be around 6\% to 8\%, depending on the finger position. This indicates that the core point is stable enough to be used as a reference point for various applications. 

Both of those findings are especially interesting for the field of contactless fingerprint analysis, where the perspective of the camera has to be accounted for. A non-vanishing core offset originating from an anatomical preference of the core towards one side would require a re-calibration of the role of the fingerprint core as a reference point. Also, the magnitude at which the core scatters around the central position of the segmented fingerprint image is crucial. A too unstable core position would make the core unreliable as reference point.

Furthermore, we found the non-central Fischer distribution to be the best matching distribution for the core's x and y position. We think that this finding can help improve the quality of synthetically generated fingerprints and further increase the understanding of the human fingerprint.

Finally, we found a correlation between the vertical offset of the core from the fingerprint center and the NFIQ 2 score. The NFIQ 2 scores favors core positions in rolled fingerprint recordings where the core sits close to the image center, but slightly off towards the lower image edge. Interestingly, we did not find any correlation of the NFIQ 2 score and the core offset in the horizontal direction.

\section*{Acknowledgment}

We gratefully acknowledge Dr. Schmid from the Austrian BMI for the continuous support during the recording sessions as well as the Bundesamt für Sicherheit in der Informationstechnik (BSI) and Jannis Priesnitz from Hochschule Darmstadt for their valuable discussions and support in this research. 
% \section*{References}

% Please number citations consecutively within brackets \cite{b1}. The 
% sentence punctuation follows the bracket \cite{b2}. Refer simply to the reference 
% number, as in \cite{b3}---do not use ``Ref. \cite{b3}'' or ``reference \cite{b3}'' except at 
% the beginning of a sentence: ``Reference \cite{b3} was the first $\ldots$''

% Number footnotes separately in superscripts. Place the actual footnote at 
% the bottom of the column in which it was cited. Do not put footnotes in the 
% abstract or reference list. Use letters for table footnotes.

% Unless there are six authors or more give all authors' names; do not use 
% ``et al.''. Papers that have not been published, even if they have been 
% submitted for publication, should be cited as ``unpublished'' \cite{b4}. Papers 
% that have been accepted for publication should be cited as ``in press'' \cite{b5}. 
% Capitalize only the first word in a paper title, except for proper nouns and 
% element symbols.

% For papers published in translation journals, please give the English 
% citation first, followed by the original foreign-language citation \cite{b6}.

\bibliographystyle{IEEEtran}
\bibliography{references}{}

\end{document}